\documentclass{tx}

\usepackage[most]{tcolorbox}
\usepackage{booktabs}
\usepackage{multirow}
\usepackage{colortbl}  
\usepackage{forest}
\usetikzlibrary{positioning}
\usepackage[newfloat,frozencache]{minted}
\usepackage{xspace}
\usepackage{array}
\usepackage{placeins}
\usepackage{cleveref}
\usepackage{float}
\usepackage{siunitx}

\definecolor{darkorange}{HTML}{EF6905}
\definecolor{codegray}{HTML}{524646}

\newcolumntype{C}[1]{>{\centering\arraybackslash}p{#1}}
\usepackage[font=small,skip=3pt]{caption}
\titlespacing*{\section}{0pt}{1.0em}{0.2em}
\titlespacing*{\subsection}{0pt}{0.6em}{0.1em}
\usepackage{pifont}
\definecolor{okgreen}{HTML}{22C55E}
\definecolor{nored}{HTML}{EF4444}
\newcommand{\yes}{\textcolor{okgreen}{\ding{51}}}  
\newcommand{\no}{\textcolor{nored}{\ding{55}}}     
\newcommand{\pt}{$\sim$}
\newcommand{\tm}{\textsc{TorchMorph}\xspace}

\setminted{
    fontsize=\scriptsize, frame=single, framesep=3pt, breaklines=true, breakanywhere=true,
    style=friendly, autogobble=true, xleftmargin=2pt }
\SetupFloatingEnvironment{listing}{name=Listing}

\logotop{0.5cm}
\logoleft{%
  \raisebox{-0.5\height}{\includegraphics[height=1.6cm]{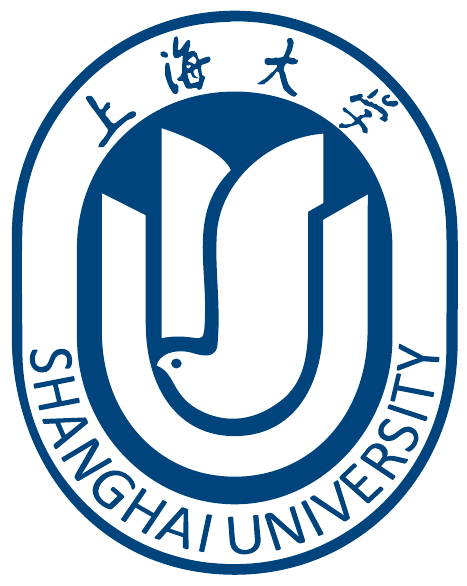}}%
  \hspace{0.35cm}%
  \raisebox{-0.5\height}{\includegraphics[height=0.78cm]{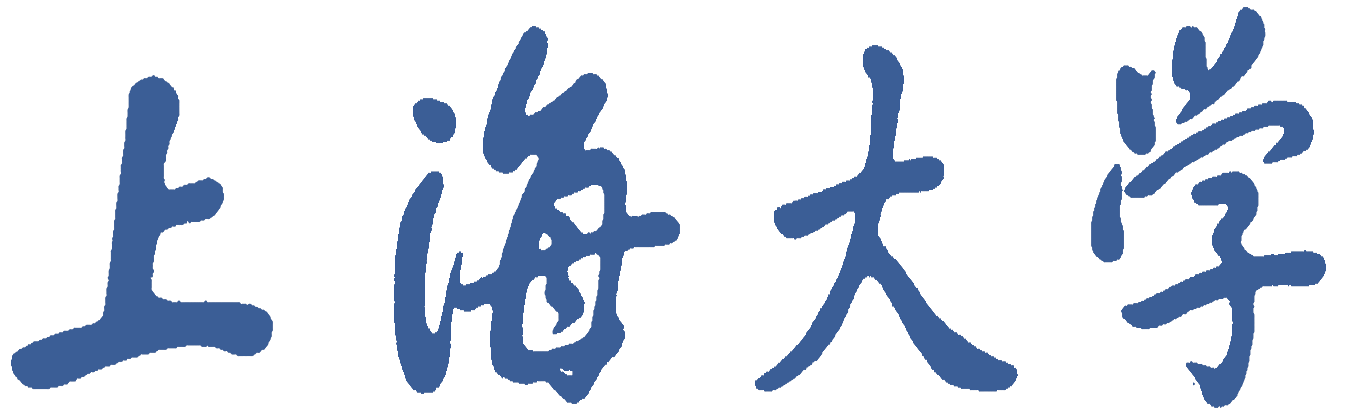}}%
}

\title{TorchMorph: CUDA-accelerated Morphological Transforms}
\author{
    Kai Zhao~\orcid{0000-0002-2496-0829} }
\affiliation{Shanghai University}
\email{kz@kaizhao.net}

\begin{abstract}
    Morphological transforms are long-standing tools for shape and mask processing,
    but the de-facto reference implementation in the Python ecosystem,
    \textit{i.e.} \texttt{scipy.ndimage},
    is CPU-only, single-array,
    and therefore unusable inside a GPU training loop without an expensive device-to-host round trip.
    GPU vision libraries built on PyTorch cover a narrow subset of these operators,
    typically restricted to two spatial dimensions and flat structuring elements.
    We present \tm,
    a lightweight PyTorch extension that closes this gap. \tm exposes 22 public operators covering binary morphology,
    greyscale morphology, exact and approximate distance transforms, and entropy-regularised optimal transport,
    all implemented as fused CUDA kernels that operate directly on \texttt{(B,\,C,\,Spatial\ldots)} CUDA tensors with up to eight spatial dimensions.
    The API deliberately mirrors \texttt{scipy.ndimage} argument-for-argument, including border modes,
    structuring-element origins and pre-allocated outputs, so that existing pipelines port with a change of import.
    We describe the layered architecture and the kernel designs behind each operator family.
    Against single-threaded CPU references, batched execution reaches up to $1.1{\times}10^{3}$ times the throughput of
    \texttt{scipy.ndimage} on greyscale morphology and up to $350\times$ on exact Euclidean distance transforms,
    while the Sinkhorn solver runs up to $42\times$ faster than POT.
    Binary and chamfer operators reproduce their SciPy counterparts exactly, and every float-valued operator agrees
    with the CPU reference to within $1.8{\times}10^{-6}$ absolute error.
    \tm is released under the MIT licence
    at \url{https://intcomp.github.io/tm}.
\end{abstract}


\begin{document}
\maketitle

{
\renewcommand\thefootnote{}
\footnotetext{
    Source code: \url{https://github.com/intcomp/torchmorph} } }

\section{Introduction}

Mathematical morphology supplies some of the oldest and most widely used primitives in image
analysis~\cite{serra1982image,soille2004morphological}.
Erosion, dilation and their compositions underpin denoising, shape decomposition and connected-component post-processing,
while the distance transform, an erosion by a parabolic structuring function, underpins skeletonisation and watershed seeding~\cite{rosenfeld1966sequential,borgefors1986distance},
boundary-aware losses~\cite{kervadec2021boundary} and Hausdorff-distance surrogates~\cite{karimi2020reducing}.
In the Python ecosystem these operators are canonically provided by \texttt{scipy.ndimage}~\cite{virtanen2020scipy},
whose semantics for border modes, structuring-element origins and connectivity conventions
have become the de-facto standard that downstream libraries are expected to reproduce.

That reference implementation was designed for a world in which images live in host memory as NumPy arrays~\cite{harris2020array}.
In modern AI-native imaging pipelines, images are represented differently in three specific ways.
First, data live on the GPU as PyTorch tensors~\cite{paszke2019pytorch},
so every call into \texttt{scipy.ndimage} forces a device-to-host copy, a single-threaded CPU computation,
and a copy back.
Second, data arrive in \emph{batches}: a training step processes tens of volumes at once,
and a per-image Python loop over a CPU routine serialises what is naturally an embarrassingly parallel workload.
Third, the interesting regime is often three- or four-dimensional (volumetric time series, multi-channel tomography),
where the CPU cost of a morphological sweep grows with the product of all spatial extents.

The obvious response is to use a GPU vision library, but that runs into a coverage problem.
Kornia~\cite{riba2020kornia} provides differentiable 2-D morphology, but not $N$-dimensional operators,
the full \texttt{scipy.ndimage} border-mode matrix, or an exact Euclidean distance transform;
cuCIM~\cite{cucim2022} offers GPU morphology through CuPy,
an interoperability layer rather than native \texttt{torch.Tensor} operators;
MONAI~\cite{cardoso2022monai} delegates several morphological post-processing steps back to SciPy or CuPy;
and OpenCV~\cite{bradski2000opencv} and scikit-image~\cite{vanderwalt2014scikit} are host-side and two-dimensional in their fast paths.
Entropic optimal transport,
increasingly used as a shape- and histogram-comparison loss~\cite{cuturi2013sinkhorn,peyre2019computational},
lives in yet another stack~\cite{flamary2021pot,feydy2019interpolating}.
A practitioner who wants batched GPU morphology,
an exact distance transform and a differentiable transport distance must therefore assemble three libraries with three tensor conventions.
\cref{tab:coverage} summarises the resulting coverage gap.

\begin{table}[!t]
    \centering
    \footnotesize
    \setlength{\tabcolsep}{3pt}
    \renewcommand{\arraystretch}{1.1}
    \begin{tabular}{@{}l ccccccc@{}}
        \toprule
        & GPU & Torch & Batch & $N$-D & Sem. & EDT & OT \\
        \midrule
        \texttt{scipy.ndimage}~\cite{virtanen2020scipy} & \no  & \no  & \no  & \yes & \yes  & \yes & \no  \\
        scikit-image~\cite{vanderwalt2014scikit}         & \no  & \no  & \no  & \yes & \pt  & \yes & \no  \\
        OpenCV~\cite{bradski2000opencv}                  & \pt  & \no  & \no  & \no  & \no  & \yes & \no  \\
        Kornia~\cite{riba2020kornia}                     & \yes & \yes & \yes & \no  & \no  & \pt  & \no  \\
        cuCIM~\cite{cucim2022}                           & \yes & \no  & \no  & \no  & \pt  & \yes & \no  \\
        MONAI~\cite{cardoso2022monai}                    & \yes & \yes & \pt  & \no  & \no  & \pt  & \no  \\
        POT~\cite{flamary2021pot}                        & \yes & \yes & \yes & \no  & \no  & \no  & \yes \\
        \midrule
        \textsc{TorchMorph}                              & \yes & \yes & \yes & \yes & \yes & \yes & \yes \\
        \bottomrule
    \end{tabular}
    \caption{
        Coverage of the capabilities \tm targets.
        \yes{}~= supported, \pt{}~= partial or indirect, \no{}~= not supported.
        \emph{Torch}~= operates on \texttt{torch.Tensor} natively;
        \emph{Batch}~= a single call over a leading batch dimension rather than a Python loop;
        \emph{$N$-D}~= more than three spatial dimensions;
        \emph{Sem.}~= matches \texttt{scipy.ndimage} border modes,
        structuring-element origins and iteration conventions;
        \emph{EDT}~= exact Euclidean distance transform.
    }
    \label{tab:coverage}
\end{table}

\tm closes that gap in one place.
It brings these operators into the PyTorch tensor world as native, batch-parallel CUDA kernels:
22 public operators spanning binary and greyscale morphology, exact and approximate distance transforms,
and entropy-regularised optimal transport.
Every one accepts \texttt{(B,\,C,\,Spatial\ldots)} CUDA tensors of spatial rank up to eight,
and mirrors its \texttt{scipy.ndimage} counterpart in name, argument order,
defaults and boundary behaviour, so porting is a change of import rather than a rewrite,
and 3-D and 4-D data are first-class rather than special cases.
The operators are fused rather than assembled from generic tensor primitives: one launch per call,
with structuring-element geometry resolved on the host, an interior fast path that removes per-axis boundary tests,
and CUDA-graph replay for the transport iterations.
The library depends only on PyTorch and a CUDA toolchain, is validated element-wise against SciPy and POT,
and is released under the MIT licence. \cref{sec:design} describes the transforms it covers and the design behind them;
\cref{sec:eval} reports its numerical agreement and throughput against the CPU reference.

\section{TorchMorph}
\label{sec:design}

\subsection{Covered transforms}
\label{sec:ops}

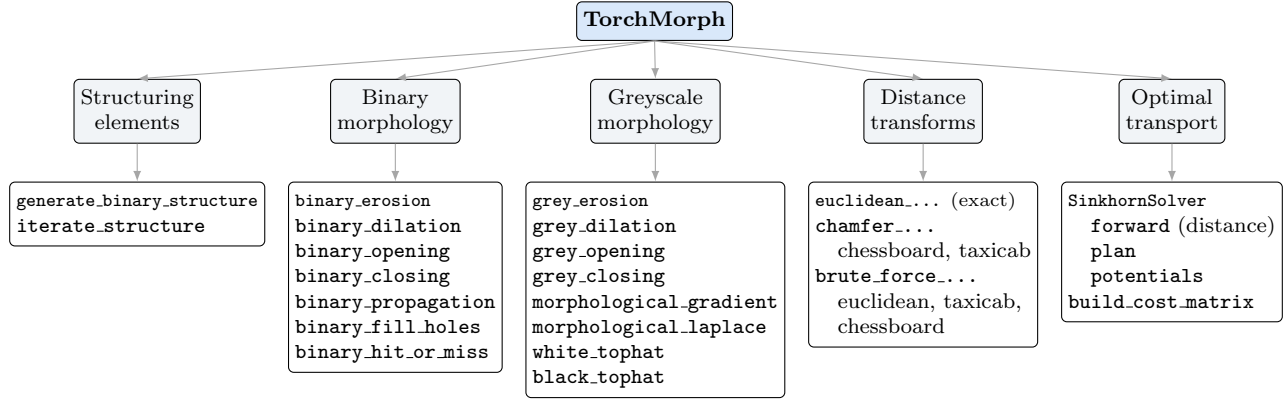
\begin{figure*}[!t]
    \centering
    \begin{forest}
        for tree={
            draw, rounded corners=2pt, align=center, font=\footnotesize,
            edge={-latex, gray!70}, parent anchor=south, child anchor=north,
            l sep=5mm, s sep=3mm, inner sep=2.5pt,
        }
        [\textbf{\tm}, fill=metablue!15
            [Structuring\\elements, fill=absbg
                [{\scriptsize
                  \texttt{generate\_binary\_structure}\\
                  \texttt{iterate\_structure}}, align=left]
            ]
            [Binary\\morphology, fill=absbg
                [{\scriptsize
                  \textbf{\texttt{binary\_erosion}}\\
                  \textbf{\texttt{binary\_dilation}}\\
                  \texttt{binary\_opening}\\
                  \texttt{binary\_closing}\\
                  \texttt{binary\_propagation}\\
                  \texttt{binary\_fill\_holes}\\
                  \texttt{binary\_hit\_or\_miss}}, align=left]
            ]
            [Greyscale\\morphology, fill=absbg
                [{\scriptsize
                  \textbf{\texttt{grey\_erosion}}\\
                  \textbf{\texttt{grey\_dilation}}\\
                  \texttt{grey\_opening}\\
                  \texttt{grey\_closing}\\
                  \texttt{morphological\_gradient}\\
                  \texttt{morphological\_laplace}\\
                  \texttt{white\_tophat}\\
                  \texttt{black\_tophat}}, align=left]
            ]
            [Distance\\transforms, fill=absbg
                [{\scriptsize
                  \textbf{\texttt{euclidean\_...}} (exact)\\
                  \textbf{\texttt{chamfer\_...}}\\
                  \quad chessboard, taxicab\\
                  \textbf{\texttt{brute\_force\_...}}\\
                  \quad euclidean, taxicab,\\
                  \quad chessboard}, align=left]
            ]
            [Optimal\\transport, fill=absbg
                [{\scriptsize
                  \textbf{\texttt{SinkhornSolver}}\\
                  \quad \texttt{forward} (distance)\\
                  \quad \texttt{plan}\\
                  \quad \texttt{potentials}\\
                  \texttt{build\_cost\_matrix}}, align=left]
            ]
        ]
    \end{forest}
    \caption{Operator taxonomy of \tm. Bold entries are backed by a dedicated
    CUDA kernel; the remainder are host-side compositions of those primitives.
    All image operators accept \texttt{(B,\,C,\,Spatial\ldots)} CUDA tensors
    with up to eight spatial dimensions.}
    \label{fig:taxonomy}
\end{figure*}

\cref{fig:taxonomy} organises the 22 exported operators into four families plus a structuring-element utility group.
Only the bold entries are backed by a dedicated CUDA kernel; the rest are host-side compositions of those primitives,
which is why adding a new top-hat variant costs no device code.

\paragraph{Binary and greyscale morphology.}
Erosion and dilation, and the openings, closings, gradients,
Laplacians and top-hats built from them~\cite{serra1982image,soille2004morphological,haralick1987image},
remain the standard tools for cleaning up masks, decomposing shapes and extracting structure at a chosen scale.
In deep-learning pipelines they appear as post-processing on predicted segmentations and as label
preparation upstream of training~\cite{cardoso2022monai,vincent1993morphological}. Both run per-step,
on batched tensors that already live on the GPU, which is exactly the regime a CPU reference serves badly. \tm implements erosion and dilation as fused kernels and derives the remaining eleven operators from them.

\paragraph{Distance transforms.}
The distance transform labels every foreground element with its distance to the nearest background element.
It underpins skeletonisation, watershed seeding and shape
descriptors~\cite{rosenfeld1966sequential,borgefors1986distance}, and in learning pipelines it is
the ingredient behind boundary-aware and Hausdorff-style
losses~\cite{kervadec2021boundary,karimi2020reducing}, which recompute a distance field every
training step and so put it on the critical path. \tm provides the exact Euclidean transform via the separable lower-envelope algorithm~\cite{felzenszwalb2012distance,maurer2003linear},
chamfer transforms under the chessboard and taxicab metrics~\cite{borgefors1986distance},
and a brute-force transform used as an exactness oracle.

\paragraph{Entropy-regularised optimal transport.}
Optimal transport compares two distributions by the cheapest way of morphing one into the other,
and its entropy-regularised form is solved by the Sinkhorn iteration~\cite{sinkhorn1967concerning,cuturi2013sinkhorn,peyre2019computational}.
It has become a standard geometry-aware loss for histograms, point clouds and attention-style matching problems,
where a bin-wise divergence would ignore how far apart the bins are. \tm provides a batched Sinkhorn solver in both the scaling and log domains,
differentiable with respect to both marginals, so it drops into a training loop as a loss.

\subsection{Architecture}
\label{sec:arch}

\subsubsection{API surface}
Every operator that consumes an image accepts a CUDA tensor shaped \texttt{(B,\,C,\,Spatial\ldots)} with $1\le\text{spatial rank}\le 8$,
validated by a single shared routine, and takes the same keyword arguments as its SciPy counterpart: \texttt{size},
\texttt{footprint}, \texttt{structure}, \texttt{mode}, \texttt{cval} and \texttt{origin} for greyscale operators;
\texttt{structure}, \texttt{iterations}, \texttt{mask} and \texttt{border\_value} for binary ones; \texttt{sampling},
\texttt{metric} and the \texttt{return\_distances}\,/\,\texttt{return\_indices} pair for distance transforms,
with pre-allocated output buffers supported throughout.
Iteration semantics match too: \texttt{iterations}$\,<1$ means \emph{iterate until the result stops changing},
which is how \texttt{binary\_propagation} and \texttt{binary\_fill\_holes} are built.

\begin{minted}{python}
# before: CPU, one volume at a time
import scipy.ndimage as ndi
out = [ndi.grey_opening(v, size=3) for v in vols]

# after: GPU, whole batch, one launch
import torchmorph as tm
out = tm.grey_opening(vols_cuda, size=3)
\end{minted}
{\captionof{listing}{Porting a SciPy pipeline.}\label{lst:port}}

\subsubsection{Three layers}
\begin{figure}[!t]
    \centering
    \begin{tikzpicture}[font=\footnotesize]
        \tikzset{
            layer/.style={draw, rounded corners=2pt, text width=6.6cm,
                          align=center, fill=absbg, inner sep=4pt},
        }
        \node[layer] (py) at (0,0)
            {\textbf{Python layer}\\[1pt]
             {\scriptsize validation $\cdot$ element resolution\\
              origin/mode normalisation $\cdot$ composition}};
        \node[layer, below=3.5mm of py] (bind)
            {\textbf{Binding layer} (\texttt{pybind11})\\[1pt]
             {\scriptsize 8 entry points $\cdot$ dtype and device guards}};
        \node[layer, below=3.5mm of bind] (cuda)
            {\textbf{CUDA layer} (6 translation units)\\[1pt]
             {\scriptsize fused morphology $\cdot$ separable EDT\\
              chamfer sweeps $\cdot$ brute force $\cdot$ Sinkhorn}};
        \node[layer, below=3.5mm of cuda, fill=white] (torch)
            {\textbf{PyTorch runtime}\\[1pt]
             {\scriptsize caching allocator $\cdot$ streams $\cdot$ CUDA graphs}};
        \draw[-latex, gray] (py) -- (bind);
        \draw[-latex, gray] (bind) -- (cuda);
        \draw[-latex, gray] (cuda) -- (torch);
    \end{tikzpicture}
    \caption{The three implementation layers. SciPy-compatible argument
    conventions are confined to the Python layer; the CUDA layer sees only
    normalised geometry.}
    \label{fig:stack}
\end{figure}
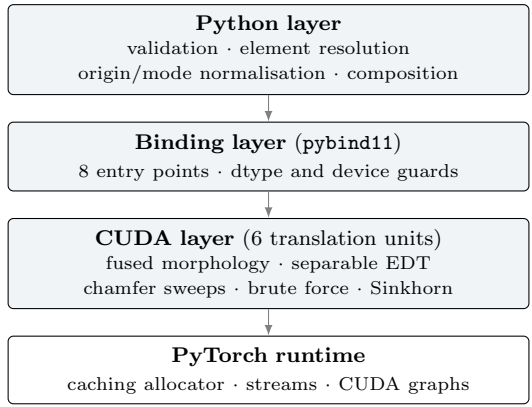

The library is organised as in \cref{fig:stack}.
The \emph{Python layer} normalises and validates arguments:
it resolves the structuring element from the \texttt{structure}\,$>$\,\texttt{footprint}\,$>$\,\texttt{size} priority chain,
expands \texttt{origin} to a per-axis tuple and range-checks it against the element extent,
maps border-mode strings to integer codes, and composes the derived operators from the two primitives.
This layer is where SciPy compatibility is defined,
and deliberately the only place that knows about argument conventions.
The \emph{binding layer} is a single \texttt{pybind11} module exposing eight kernel entry points compiled from six CUDA translation units;
keeping that surface small means derived operators cost no extra device code.
The \emph{kernel layer} holds the CUDA implementations, with two decisions recurring across it:
all geometry that can be resolved on the host is resolved on the host,
and every kernel is written against a runtime spatial rank with a compile-time bound,
so coordinate scratch space stays in registers; the Python layer caps that rank at eight.

\subsection{Kernel design}
\label{sec:kernels}

\subsubsection{Fused morphology kernel}
A naive GPU morphology kernel recomputes, for every output element and every structuring-element position,
a full $N$-dimensional coordinate mapping with a per-axis boundary test.
For a $3^3$ element in 3-D that is 27 mapped coordinates per voxel, each requiring the linear index
to be decomposed along every axis before it can be bounds-checked.

\tm avoids this in two ways. On the host, the structuring element is flattened once into a list of active entries;
for each we store the per-axis offset \emph{and} the precomputed flat offset against the input's spatial strides,
and inactive footprint positions never reach the device.

On the device, each thread first tests whether its output coordinate is \emph{interior},
using per-axis offset extrema computed on the host.
Interior threads, the overwhelming majority, take a fast path that adds the precomputed flat offset directly to the linear index,
with no per-axis arithmetic and no boundary test at all.
Only threads within one element radius of a face take the general path,
which resolves out-of-bounds neighbours as its SciPy counterpart does:
the greyscale kernel implements all five border modes (\texttt{constant}, \texttt{reflect},
\texttt{nearest}, \texttt{mirror}, \texttt{wrap}), the binary kernel a single
\texttt{border\_value}.

Erosion and dilation are instantiations of one templated kernel parameterised by a functor supplying the identity element and the combining rule,
and the binary kernel adds a \texttt{done} predicate so a thread can break out of the reduction as soon as the result is decided,
false for erosion and true for dilation, which is a substantial saving on sparse masks.

\subsubsection{Distance-transform kernels}
The EDT uses the separable lower-envelope formulation,
in which each one-dimensional pass computes the lower envelope of a family of parabolas in a single sweep~\cite{felzenszwalb2012distance}.
That sweep is inherently sequential, so \tm assigns one thread block per scanline:
a single thread builds the envelope in shared memory while the whole block cooperates on loading the line and,
afterwards, on the query phase, where each output position binary-searches the intersection array in parallel.
Parallelism comes from the number of scanlines, which is ample:
for $B$ batch items of $C$ channels and extent $n^d$ it is $B\,C\,n^{d-1}$ per pass.

Three paths exist:
a 2-D specialisation for extents up to $2048$ that keeps the row pass fully contiguous and fuses the final square root into the column pass;
a general path that transposes the active axis to be innermost so every pass sees contiguous memory;
and a fallback that spills the envelope stack to a lazily allocated global buffer when a scanline exceeds the shared-memory budget.
Nearest-background indices are propagated through the passes on request,
and an image with no background is handled by the same virtual out-of-bounds convention SciPy uses.

The chamfer transform is implemented as dimension-separable forward and backward sweeps,
with extra diagonal passes for the chessboard metric.
The brute-force transform stages background coordinates through shared memory in tiles of 256 and is templated over both metric and spatial rank,
so the coordinate loop is fully unrolled and metric selection costs no branch;
its quadratic cost confines it to validating the separable transforms and to the one case they miss:
the chamfer entry point takes no \texttt{sampling} argument, so anisotropic chessboard and taxicab
distances are reachable only this way.

\subsubsection{Batch-tiled Sinkhorn solver}
\label{sec:ot}
The transport module solves the Sinkhorn iteration for a batch of $n$ histogram pairs of dimension $d$ sharing one $d\times d$ cost matrix,
and three characteristics of that workload drive the design.

\emph{The matrix is shared across the batch}:
rather than the generic batched matrix--vector product, which reads the $d^2$ matrix once per batch item,
\tm assigns one block per (row, batch-tile) pair with a tile of eight items,
streams the matrix row once and applies it to all eight scaling vectors held in registers,
dropping matrix traffic by the tile factor.

\emph{The log-domain update needs only one pass}:
where the textbook log-sum-exp makes two passes over the row, one for the maximum and one for the shifted sum,
\tm maintains a running maximum and a rescaled running sum in a single pass and reduces the resulting pairs across the block with a merge operator that is well defined on the empty state,
so an all-zero marginal pins the potential to $-\infty$ instead of producing \texttt{NaN}.

\emph{Long runs are launch-latency bound}:
each iteration is two small kernels,
so from 100 requested iterations upwards \tm captures a chunk of 25 into a CUDA graph~\cite{nvidia2024cuda} and replays it.
Because the iterations are fixed-point steps on ping-pong buffers, executing extra iterations is always safe,
and the graph path degrades gracefully to plain launches when capture is unavailable.

Gradients come from a custom autograd function returning the centered dual potentials,
which by the envelope theorem are the exact gradients of the entropic transport cost with respect to the marginals~\cite{peyre2019computational};
the potentials are centred and stashed during the forward pass, so the backward pass is a single
broadcast multiply and never differentiates through the iteration.

\subsection{Testing and reference alignment}
\label{sec:testing}
Matching \texttt{scipy.ndimage} is a claim about behaviour, so it is checked by differential
testing rather than asserted. The suite is 78 test functions across five modules.

\emph{Oracle agreement} accounts for most of it: every exported operator is compared elementwise
against its reference (\texttt{scipy.ndimage} for morphology and distance transforms, POT for
transport) over 2-D, 3-D and higher-rank inputs, batch and channel combinations, non-contiguous
and transposed layouts, anisotropic sampling, every border mode, and asymmetric structuring
elements with shifted origins. The reference is applied sample by sample to the same array the GPU
receives, so batching cannot mask a per-item discrepancy. Where no reference settles the question,
\emph{invariants} take over: transport plans must reproduce both marginals, returned indices must
point at a genuine nearest background element, and the analytic gradient is checked against a
finite-difference directional derivative.

The remainder guards the seams. \emph{Cross-implementation} tests solve one problem along every
path the library can take: fused kernels against the pure-torch fallback, float32 against float64,
CPU against GPU, CUDA-graph replay against plain launches, early-stopped against fully iterated.
The results must coincide. \emph{Contract} tests, a quarter of the suite,
assert failure on spatial rank above eight, mismatched origin, mask or output shapes, unknown modes
and metrics, and non-CUDA input. \emph{Runtime} tests issue work on a side stream and a non-default
device, checking that kernels honour the caller's stream and device rather than the defaults.

\section{Results}
\label{sec:eval}

\paragraph{Protocol.}
All comparisons use \texttt{scipy.ndimage} as the reference, except for transport, which uses POT~\cite{flamary2021pot}.
All measurements come from a single machine.
\emph{Hardware}: NVIDIA GeForce RTX 4090 D GPU (Ada Lovelace, 48 GB, driver 580.173.02),
2 × Intel Xeon Gold 6330 CPU (56~cores / 112~threads, 2.00 GHz),
and 128 GB of system memory.
\emph{Software}: Ubuntu 24.04.3 LTS, CUDA 12.4, Python 3.12.13, PyTorch 2.6.0+cu124,
NumPy 2.5.1, SciPy 1.18.0, and POT 0.9.6.post1.
\textsc{TorchMorph} extensions were compiled with \texttt{-O3} and \texttt{--use\_fast\_math}.
The SciPy and POT baselines are single-threaded and therefore occupy one CPU core,
while every GPU result uses a single device.
Accuracy is measured by transferring the GPU result to the host and comparing against the reference output on the identical input;
we report the maximum absolute error $\max|y-\hat y|$ and the relative error $\lVert y-\hat y\rVert_2/\lVert y\rVert_2$.
Timings are collected with \texttt{torch.utils.benchmark},
taking the median of blocked auto-ranged measurements with a one-second minimum run time and per-item normalisation,
after warm-up and with an explicit \texttt{torch.cuda.synchronize()} inside the timed region.
Because the point of the library is batching, we report two GPU columns: \emph{$1\times$},
a Python loop calling the operator once per batch item, the honest analogue of a SciPy loop, and \emph{batch},
a single call on the whole batch.
Speed-up is SciPy time divided by batched GPU time per item.
Every table is reproduced by the scripts in \href{https://github.com/intcomp/torchmorph/tree/main/benchmark}{benchmark/}.

\begin{table}[!t]
    \centering
    \footnotesize
    \setlength{\tabcolsep}{4pt}
    \begin{tabular}{llcc}
        \toprule
        Operator family & Reference & Max abs.\ err. & Rel.\ $\ell_2$ err. \\
        \midrule
        Binary morphology      & \texttt{ndimage} & $0$ & $0$ \\
        Grey morphology        & \texttt{ndimage} & $4.77{\times}10^{-7}$ & $2.24{\times}10^{-8}$ \\
        Euclidean DT           & \texttt{ndimage} & $2.06{\times}10^{-7}$ & $6.77{\times}10^{-9}$ \\
        Chamfer DT (chessb.)   & \texttt{ndimage} & $0$ & $0$ \\
        Chamfer DT (taxicab)   & \texttt{ndimage} & $0$ & $0$ \\
        Brute-force DT ($\ell_2$)
                               & \texttt{ndimage} & $2.06{\times}10^{-7}$ & $6.71{\times}10^{-9}$ \\
        Sinkhorn distance      & POT              & $1.75{\times}10^{-6}$ & $8.82{\times}10^{-8}$ \\
        \bottomrule
    \end{tabular}
    \caption{Worst-case numerical error over all tested
    configurations against the CPU reference implementations.
    Binary morphology and chamfer distance transforms match
    \texttt{scipy.ndimage} exactly in the tested cases, while
    float-valued operators differ only at float32-level numerical
    precision. Sinkhorn distance is compared against POT.
    }
    \label{tab:accuracy}
\end{table}

\paragraph{Numerical agreement.}
Table~\ref{tab:accuracy} shows close agreement with the CPU references.
Binary morphology and chamfer distance transforms match \texttt{scipy.ndimage} exactly;
all float-valued operators have worst-case absolute and relative $\ell_2$ errors below $1.8{\times}10^{-6}$ and $9{\times}10^{-8}$,
respectively.
Small discrepancies arise from float32 GPU arithmetic and \texttt{--use\_fast\_math}.
NaN propagation is the only documented behavioral difference from SciPy.
Correctness is covered by 78 test functions, parametrised over operator, spatial rank,
input shape, origin, sampling, requested outputs, device and solver variant, with
\texttt{scipy.ndimage} as the oracle for morphology and distance transforms and
POT for transport.
The suite runs on every push and pull request through a self-hosted continuous-integration
runner equipped with a physical CUDA device, so each change is checked against the CPU
references by executing the actual kernels rather than a mocked or CPU-only code path;
a second hosted runner enforces formatting and static checks.

\paragraph{Throughput.}
\Cref{tab:throughput} reports per-input throughput for morphology and distance transforms.
The $B=1$ results capture single-input GPU execution, where launch overhead can dominate,
while larger batches expose the parallel regime targeted by \textsc{TorchMorph}.
The batching benefit generally decreases as individual inputs become large enough to better utilize the GPU.

\begin{table}[!t]
    \centering
    \caption{
    Per-input throughput across representative operators, input sizes,
    and batch sizes. Throughput is reported in inputs/ms
    (higher is better), where one input denotes one 2-D image or one
    3-D volume. {TM} = \textsc{TorchMorph}; its values are rounded to one decimal place, while the SciPy
    column keeps three, since one decimal would collapse most of its entries
    to 0.0.
    }
    \label{tab:throughput}

    \footnotesize
    \renewcommand{\arraystretch}{1.05}
    \setlength{\tabcolsep}{2pt}
    \setlength{\arrayrulewidth}{0.4pt}

    \begin{tabular}{
        @{}
        l
        C{0.76cm}
        C{1.08cm}
        *{4}{C{0.76cm}}
        @{}
    }
        \toprule

        \multirow{3}{*}{Operator}
        & \multirow{3}{*}{Size}
        & \multicolumn{5}{c}{Throughput (inputs/ms) $\uparrow$} \\

        \cmidrule(lr){3-7}

        & &
        \multirow{2}{*}{SciPy}
        & \multicolumn{4}{c}{TM} \\

        \cmidrule(lr){4-7}

        & & & $B{=}1$ & $B{=}2$ & $B{=}4$ & $B{=}8$ \\

        \midrule

        & $256^2$
        & 0.865
        & 6.9
        & 13.7
        & 27.8
        & 55.6 \\

        \multirow{-2}{*}{Binary erosion} & $1024^2$
        & 0.056
        & 6.8
        & 12.8
        & 22.2
        & 35.7 \\

        \addlinespace[1pt]

        \rowcolor{gray!18}
        & $256^2$
        & 0.851
        & 6.7
        & 13.5
        & 27.0
        & 52.6 \\

        \rowcolor{gray!18}
        \multirow{-2}{*}{Binary dilation} & $1024^2$
        & 0.057
        & 6.7
        & 12.5
        & 21.7
        & 35.7 \\

        \addlinespace[1pt]

        & $256^2$
        & 0.740
        & 16.1
        & 32.3
        & 62.5
        & 83.3 \\

        \multirow{-2}{*}{Grey erosion} & $1024^2$
        & 0.040
        & 13.7
        & 23.3
        & 34.5
        & 45.5 \\

        \addlinespace[1pt]

        \rowcolor{gray!18}
        & $256^2$
        & 0.756
        & 10.3
        & 30.3
        & 58.8
        & 111.1 \\

        \rowcolor{gray!18}
        \multirow{-2}{*}{Grey dilation} & $1024^2$
        & 0.040
        & 13.2
        & 22.2
        & 33.3
        & 45.5 \\

        \addlinespace[2pt]
        \midrule
        \addlinespace[1pt]

        & $256^2$
        & 0.160
        & 13.3
        & 25.6
        & 30.3
        & 40.0 \\

        \multirow{-2}{*}{EDT 2-D} & $1024^2$
        & 0.011
        & 2.1
        & 2.7
        & 2.8
        & 2.9 \\

        \addlinespace[1pt]

        \rowcolor{gray!18}
        & $64^3$
        & 0.031
        & 6.3
        & 8.3
        & 9.3
        & 10.8 \\

        \rowcolor{gray!18}
        \multirow{-2}{*}{EDT 3-D} & $128^3$
        & 0.003
        & 1.4
        & 1.4
        & 1.5
        & 1.5 \\

        \addlinespace[1pt]

        & $256^2$
        & 0.585
        & 4.7
        & 9.4
        & 18.2
        & 33.3 \\

        \multirow{-2}{*}{CDT chessboard} & $1024^2$
        & 0.037
        & 1.5
        & 2.6
        & 4.3
        & 5.9 \\

        \addlinespace[1pt]

        \rowcolor{gray!18}
        & $256^2$
        & 0.637
        & 5.5
        & 10.9
        & 20.0
        & 43.5 \\

        \rowcolor{gray!18}
        \multirow{-2}{*}{CDT taxicab} & $1024^2$
        & 0.041
        & 1.7
        & 3.4
        & 6.6
        & 11.6 \\

        \addlinespace[1pt]

        BFDT ($\ell_2$)
        & $256^2$
        & \mbox{$<0.001$}
        & 1.1
        & 1.2
        & 1.2
        & 1.2 \\

        \bottomrule
    \end{tabular}
\end{table}

\paragraph{Morphology and distance transforms.}
\Cref{tab:throughput} reports per-input throughput for representative morphological operators and distance transforms.
Across most operators, throughput increases substantially with batch size,
reflecting the fact that a single input does not supply enough parallel work to saturate the GPU.
Grey dilation on $256^2$ images, for instance,
improves from $10.31$ inputs/ms at $B=1$ to $111.11$ inputs/ms at $B=8$ (a $10.8\times$ gain),
while grey erosion improves from $16.13$ to $83.33$ inputs/ms ($5.2\times$).
Binary morphology follows the same pattern,
reaching $36$--$56$ inputs/ms at $B=8$ depending on the operator and image size.

The benefit of batching diminishes markedly for computationally heavier transforms,
where a single large input already supplies enough parallelism to approach device saturation on its own.
EDT on $1024^2$ images improves only modestly, from $2.11$ to $2.86$ inputs/ms ($1.4\times$),
and EDT on $128^3$ volumes remains essentially flat at approximately $1.4$ inputs/ms regardless of batch size.
CDT retains a clear batching benefit, particularly for smaller inputs, whereas BFDT shows only a marginal gain,
from $1.15$ to $1.25$ inputs/ms.
SciPy's throughput on BFDT is extremely low because it relies on a brute-force reference implementation,
so this comparison should be read as a correctness oracle rather than as a benchmark against an optimized CPU distance-transform implementation.

\begin{table}[!t]
    \centering
    \caption{
    Entropic optimal transport against POT on 2-D grid problems with
    $d=32^2$ bins ({TM} = \textsc{TorchMorph}). The 1000-iteration row uses
    CUDA-graph replay. Timings are
    rounded to one decimal place; speed-ups are computed from the unrounded
    values.
    }
    \label{tab:ot}

    \footnotesize
    \setlength{\tabcolsep}{2pt}
    \renewcommand{\arraystretch}{0.96}

    \begin{tabular}{@{}l S[table-format=3.1] S[table-format=2.1] cc@{}}
        \toprule
        Configuration
        & {POT}
        & {TM}
        & Speed-up
        & Rel.\ err. \\
        & {(ms)}
        & {(ms)}
        &
        & (plan) \\
        \midrule

        scaling, 100 it.
        & 23.0
        & 1.2
        & $18.8\times$
        & $9.21{\times}10^{-4}$ \\

        scaling, 1000 it.
        & 229.7
        & 10.1
        & $22.8\times$
        & $4.45{\times}10^{-7}$ \\

        log-domain, 200 it.
        & 35.4
        & 3.0
        & $11.8\times$
        & $4.55{\times}10^{-5}$ \\

        batch $n=16$, 100 it.
        & 367.3
        & 8.7
        & $42.4\times$
        & $1.54{\times}10^{-3}$ \\

        \bottomrule
    \end{tabular}
\end{table}

\paragraph{Optimal transport.}
\Cref{tab:ot} compares the Sinkhorn solver against POT~\cite{flamary2021pot} on grid-structured problems,
where the cost matrix is the pairwise $\ell_2$ distance between the points of a $32\times32$ grid.
We also report the relative error of the recovered transport plan.
Three configurations are of particular interest: the scaling form at moderate regularisation,
the log-domain form at small regularisation where the scaling form underflows,
and a batched run that exercises the tiling described in \cref{sec:ot}.

\paragraph{Threats to the comparison.}
The speed-ups therefore compare a GPU against one CPU core, not a well-parallelised CPU implementation;
we report per-item times and the $B=1$ column so the batching effect can be separated from the device effect.
The brute-force timings are the cost of a correctness oracle, not a performance claim.

\section{Conclusion}

\tm provides batch-parallel, $N$-dimensional CUDA implementations of binary and greyscale
morphology, exact and approximate distance transforms, and entropy-regularised optimal transport,
behind an API that mirrors \texttt{scipy.ndimage}. Its contribution is availability rather than
algorithmic novelty: classical algorithms implemented once, correctly, against the tensor
conventions modern imaging pipelines actually use, and validated against the references the
community already trusts. That turns these operators from an offline preprocessing step into
something that can sit inside a training loop~\cite{kervadec2021boundary,karimi2020reducing}.

Three limitations remain. The morphology and distance kernels are forward-only; only the transport
module is differentiable, though erosion and dilation admit subgradients routed to the
arg-min/arg-max position. They also require CUDA, the intended fallback being SciPy itself, whereas
the transport module drops back to pure \texttt{torch} on CPU. And they compute in float32 under
\texttt{--use\_fast\_math}, with \texttt{NaN} propagation not guaranteed to match the reference.
Future work is autograd for the morphological operators and a wider set of connected-component and
reconstruction operators.

\section*{Acknowledgements}

\let\oldthebibliography\thebibliography
\renewcommand{\thebibliography}[1]{%
  \oldthebibliography{#1}%
  \setlength{\itemsep}{0pt}\setlength{\parsep}{0pt}\setlength{\parskip}{0pt}}
{\scriptsize
\bibliographystyle{unsrt}
\bibliography{refs}
}

\end{document}